%% file: main.tex
\documentclass[10pt,twocolumn,letterpaper]{article}

\usepackage[pagenumbers]{iccv} 


\definecolor{iccvblue}{rgb}{0.21,0.49,0.74}
\usepackage[pagebackref,breaklinks,colorlinks,allcolors=iccvblue]{hyperref}

\usepackage{graphicx}
\usepackage{multirow}
\usepackage{caption}
\usepackage{booktabs}
\input{math_commands}

\newcommand{\mname}{GausSim}
\newcommand{\dname}{READY}
\def\Tabref#1{Table~\ref{#1}}

\def\paperID{****} 
\def\confName{ICCV}
\def\confYear{2025}

\title{\mname: Foreseeing Reality by Gaussian Simulator for Elastic Objects}

\author{
  Yidi Shao$^1$ \quad Mu Huang$^{2,4}$ \quad Chen Change Loy$^{1}$ \quad Bo Dai$^{3}$\\
$^1$S-Lab Nanyang Technological University, $^2$Fudan University\\
$^3$The University of Hong Kong, $^4$Shanghai Artificial Intelligence Laboratory\\
{\tt\small yidi001@e.ntu.edu.sg, mhuang24@m.fudan.edu.cn, ccloy@ntu.edu.sg, bdai@hku.hk}
}

\begin{document}
\input{tab_fig/0_teaser_v2}
\input{sec/0_abstract}    
\input{sec/1_intro}
\input{sec/2_related_work}
\input{sec/3_method}
\input{sec/5_experiments}
\input{sec/6_conclusion}
{
    \small
    \bibliographystyle{ieeenat_fullname}
    \bibliography{main}
}

\input{sec/X_suppl}

\end{document}

%% file: math_commands.tex
\usepackage{amsmath,amsfonts,bm}

\def\Figref#1{Figure~\ref{#1}}

\def\Secref#1{Section~\ref{#1}}

\def\eqref#1{equation~\ref{#1}}
\def\Eqref#1{Equation~\ref{#1}}

\def\1{\bm{1}}

\def\va{{\bm{a}}}

\def\vc{{\bm{c}}}
\def\vd{{\bm{d}}}

\def\vx{{\bm{x}}}

\def\mF{{\bm{F}}}

\def\mR{{\bm{R}}}

\def\mU{{\bm{U}}}
\def\mV{{\bm{V}}}

\def\mX{{\bm{X}}}

\DeclareMathAlphabet{\mathsfit}{\encodingdefault}{\sfdefault}{m}{sl}
\SetMathAlphabet{\mathsfit}{bold}{\encodingdefault}{\sfdefault}{bx}{n}



%% file: tab_fig/0_teaser_v2.tex
\twocolumn[{%
      \renewcommand\twocolumn[1][]{#1}%
      \maketitle
      \vspace{-1.3cm}
      \begin{center}
        \centering
        \includegraphics[width=0.97\textwidth]{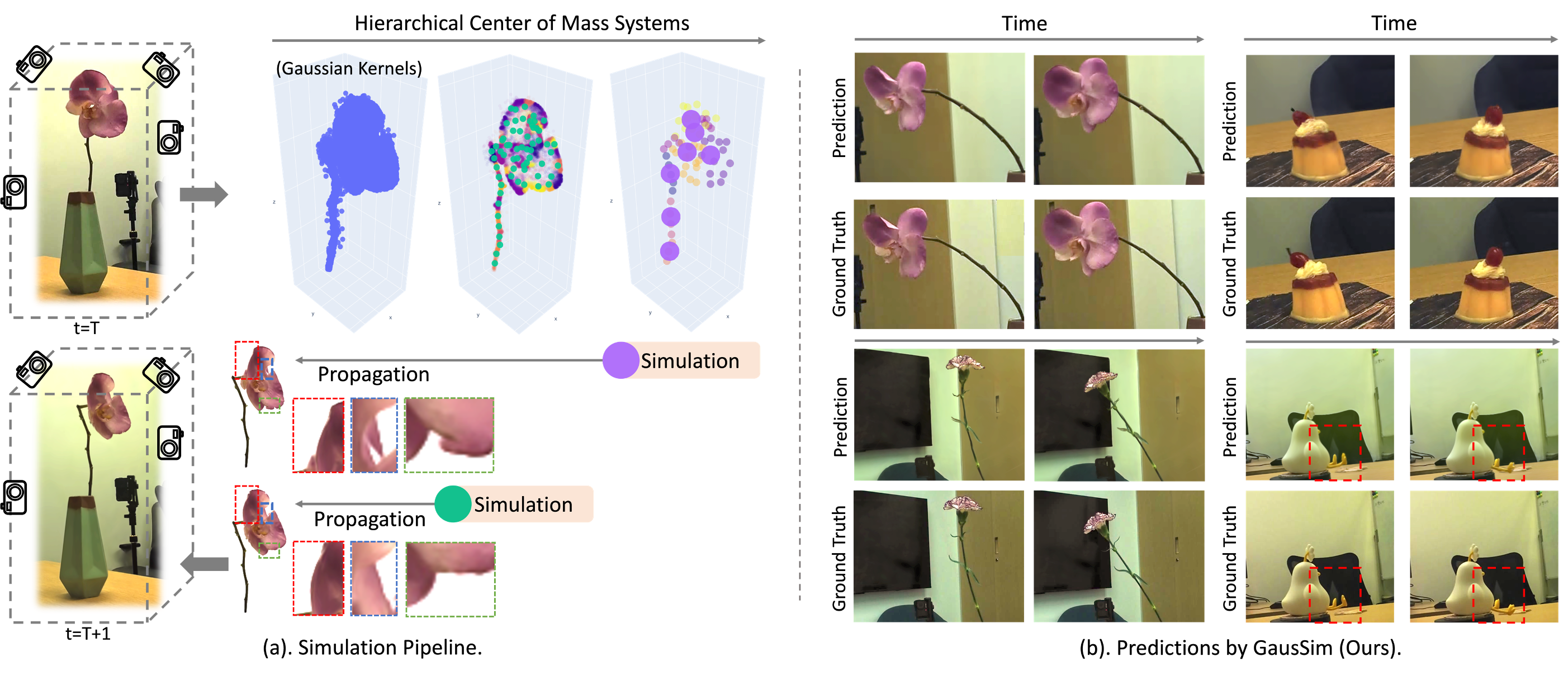}\vspace{0.1cm}
        \vskip -0.6cm
        \captionof{figure}{To bridge the gap between virtual environment and real world,
        we present \mname,
        a neural network-based physics simulator tailored for objects represented through Gaussian Splatting \cite{DBLP:journals/tog/KerblKLD23}.
        (a). \mname~is formulated based on continuum mechanics with explicit physics constraints. In addition, we propose a hierarchical structure that is simulated and propagated iteratively and enables the addition of finer details to coarser results, achieving efficient simulation (95\% reduction of kernel-wise computations) with high fidelity.
        We verify \mname~on both synthetic dataset and our REAl DYnamic dataset, called \dname, which includes complex elastic deformations.
        (b). \mname~achieves faithful and vivid predictions.
        \textit{Please refer to the supplementary video for better visualization.}
        }
        \label{fig:teaser}
      \end{center}
    }]

%% file: sec/0_abstract.tex
\begin{abstract}
We introduce \mname, a novel neural network-based simulator designed to capture the dynamic behaviors of real-world elastic objects represented through Gaussian kernels. 
We leverage continuum mechanics and treat each kernel as a Center of Mass System (CMS) that describes continuous piece of matter, accounting for realistic deformations without idealized assumptions.
To improve computational efficiency and fidelity, we employ a hierarchical structure that further organizes kernels into CMSs with explicit formulations, enabling a coarse-to-fine simulation approach. This structure significantly reduces computational overhead while preserving detailed dynamics. In addition, \mname~incorporates explicit physics constraints, such as mass and momentum conservation, ensuring interpretable results and robust, physically plausible simulations. To validate our approach, we present a new dataset, \dname, containing multi-view videos of real-world elastic deformations. Experimental results demonstrate that \mname~achieves superior performance compared to existing physics-driven baselines, offering a practical and accurate solution for simulating complex dynamic behaviors.
Code and model are available at our \href{https://www.mmlab-ntu.com/project/gausim/index.html}{project page}
\end{abstract}

%% file: sec/1_intro.tex
\section{Introduction}
\label{sec:intro}

Understanding and simulating dynamic processes is essential in fields
as diverse as ecology, climatology, physics, and computer graphics.
For instance,
by replicating the dispersal of dandelion seeds or the gentle swaying of flowers,
we can create realistic applications in animation, gaming, filmmaking, and virtual reality.
However,
accurately simulating real-world dynamics remains a challenge
due to the gap between real-world complexities and experimental models,
which often depend on idealized conditions and simplified settings.

Some methods \cite{DBLP:journals/corr/abs-2312-17142, DBLP:conf/cvpr/BahmaniSRWGWTPT24} have adopted generative models to predict the dynamics of complex objects,
such as cartoon characters.
However,
because these models rely solely on video-based supervision without explicit physics constraints,
they often fail to adhere to fundamental physics laws,
like Newton's laws,
and struggle with making accurate long-term predictions.
Other approaches \cite{DBLP:conf/cvpr/XieZQLF0J24, DBLP:journals/corr/abs-2403-09434, DBLP:journals/corr/abs-2404-13026, DBLP:journals/corr/abs-2409-05819, NeuMA_NIPS24} have integrated physics priors,
such as the Material Point Method (MPM) \cite{DBLP:conf/siggraph/JiangSTSS16},
into the reconstructed 3D Gaussian representations.
These models often treat Gaussian kernels as particles,
which serve as simulation units within the chosen analytical physics framework.
Nevertheless,
because these analytical models are derived under idealized assumptions,
they often fall short in capturing the full complexity of real-world scenarios,  limiting their ability to fully represent intrinsic physics laws.

In this paper,
we introduce \mname,
a novel neural network-based simulator designed to bring physics-based interpretability to Gaussian representations
and capture the underlying physics laws for elastic dynamics in reality.
\mname~is tailored specifically to handle objects represented through Gaussian Splatting \cite{DBLP:journals/tog/KerblKLD23},
treating each Gaussian kernel as a CMS that accounts for continuous piece of matter.
This enables \mname~to simulate the behaviors of elastic objects using principles from continuum mechanics.
Additionally,
\mname~leverages a unique hierarchical structure to more efficiently and accurately 
simulate densely distributed kernels,
capturing realistic deformations with high fidelity.
Finally,
we apply explicit physics constraints to our \mname,
enabling it to preserve properties of mass and momentum conservations.

Our \mname~benefits the simulation of Gaussian kernels in several ways.
First,
the continuum mechanism applies to most cases of elastic deformations in real life,
while the neural network-based simulator can learn to simulate realistic scenarios without relying on idealized assumptions.
By combining these two components, \mname~enhances the ability to learn the underlying physics in real-world phenomena.
Second,
the hierarchical structure acts as a coarse-to-fine approach as shown in \Figref{fig:teaser}(a),
solving kernel-wise dynamics more efficiently while
preserving high fidelity.
In addition,
since the constructed hierarchical structure is agnostic to the shape of the simulated objects,
\mname~is highly generalizable to various object types.
Third,
with explicit physics constraints,
\mname~intrinsically enforces mass conservation and preserves momentum,
leading to interpretable formulations
and robust performance.

We validate the effectiveness of our method on both real and synthetic datasets. Specifically,
we introduce a dataset collecting REAl DYnamics, called \dname,
which consists of multi-view videos capturing the elastic deformations of real-world objects, including examples like moth orchids, carnation, pudding, and duck.
Unlike video diffusion model priors focusing primarily on motion quality, our real-world dataset ensures the dynamics are grounded in real-world physics.
Our \dname~contains over 30 dynamic trajectories for each object,
with each sequence consisting of 100 frames.
The synthetic data include the dynamics of an elastic bunny.
We compare \mname~with baselines that use analytical physics priors,
such as those based on MPM simulators \cite{DBLP:conf/cvpr/XieZQLF0J24, DBLP:journals/corr/abs-2404-13026},
assessing simulation accuracy and quality against the ground truth multi-view videos.
The experiments demonstrate that \mname~achieves superior performance
and faithfully captures real-world physics laws.

Our contributions can be summarized as follows:
1) we propose \mname,
a neural network-based simulator that integrates continuum mechanics to realistically simulate dynamics without idealized conditions,
bridging the gap between simulated environment and reality;
2) we employ a hierarchical structure
to efficiently simulate the dynamics of Gaussian kernels, reducing the number of predictions by around 95\%, while maintaining high fidelity;
3) we enforce explicit physics principles, such as mass and momentum conservation, in \mname, resulting in an interpretable model and robust simulation;
4) we present \dname, a multi-view video dataset capturing dynamics of elastic deformations of real-world objects.

%% file: sec/2_related_work.tex
\section{Related Work}
\paragraph{Differentiable Simulation.}
Differentiable simulation enables the integration with workflows using gradient-based optimization, such as the inverse problems in physics and robotics.
Traditional simulations based on partial differential equations (PDEs) typically employ Material Point Method (MPM) \cite{DBLP:conf/siggraph/JiangSTSS16},
Position Based Dynamics (PBD) \cite{muller2007position}, and Finite Element Method (FEM) \cite{belytschko2014nonlinear, hughes2003finite}.
These techniques are commonly adopted in tasks including physics parameter estimation \cite{DBLP:conf/cvpr/ChenTSKKVL22, DBLP:journals/corr/abs-2406-04338},
constitutive relations \cite{DBLP:conf/icml/0002CDT0GM23},
fluid \cite{DBLP:conf/iclr/XianZXT0FG23},
animations \cite{DBLP:journals/pacmcgit/StuyckC23},
robotics \cite{DBLP:conf/rss/HeidenMNFGR21, DBLP:journals/finr/DegraveHDW19, DBLP:conf/corl/LinFWYCFWXG24},
reconstructions \cite{DBLP:conf/nips/GuoW0ZOG0HM24, DBLP:conf/iclr/XiongHL0G025}, \etc.
In addition,
neural network-based methods \cite{DBLP:journals/jcphy/RaissiPK19, DBLP:conf/iclr/PfaffFSB21, DBLP:conf/iclr/LiWTTT19, DBLP:conf/iclr/UmmenhoferPTK20, 
DBLP:conf/eccv/ShaoLD22,
DBLP:conf/iccv/ShaoL023, ShaoEUNet24} have been proposed to simulate complex dynamics,
often achieving more efficient and robust performance.
In this work,
we aim to capture the underlying physics laws by a neural network-based model,
which can handle complex scenarios without relying on idealized conditions.

\noindent\textbf{4D Generation.}
Almost everything in real life is dynamic, and much of its meaning comes from these dynamics. Capturing and generating the dynamics of 3D objects has been a significant focus in research. 
Some methods \cite{DBLP:conf/icml/SingerSPAMKGVP023, DBLP:conf/mm/ShenLSPX0L23, DBLP:journals/corr/abs-2312-17142, DBLP:conf/cvpr/Ling00FK24, DBLP:conf/eccv/BahmaniLYSRLLPTWTL24, DBLP:journals/corr/abs-2312-17225, DBLP:conf/cvpr/BahmaniSRWGWTPT24} create 4D contents using video diffusion models.
While these approaches have achieved notable progress, they often fail to adhere to the real-world laws of physics.
To produce physically plausible predictions,
recent approaches \cite{DBLP:conf/iclr/LiQCJLJG23, DBLP:conf/cvpr/XieZQLF0J24, DBLP:journals/corr/abs-2404-13026, DBLP:conf/cvpr/FengSLSJ024, DBLP:journals/corr/abs-2409-05819, DBLP:journals/corr/abs-2401-15318, DBLP:conf/siggraph/JiangYXLFWLLG0J24, DBLP:journals/corr/abs-2406-04338} introduce physics-based priors in their models.
For example,
several approaches employ physics simulation engines to guide the generation of dynamics,
such as those based on Material Point Method (MPM) \cite{DBLP:journals/corr/abs-2409-05819, DBLP:journals/corr/abs-2406-04338, DBLP:journals/corr/abs-2404-13026, NeuMA_NIPS24} and Position Based Dynamics \cite{DBLP:journals/corr/abs-2401-15318, DBLP:conf/siggraph/JiangYXLFWLLG0J24}.
To closely mimic real-world dynamics, unknown material properties or physics parameters are estimated~\cite{DBLP:journals/corr/abs-2406-04338, DBLP:journals/corr/abs-2404-13026} to better align the simulation engines with generated video data.
Besides,
some researchers use analytical material models,
such as the mass-spring system \cite{DBLP:journals/corr/abs-2403-09434},
to approximate and reconstruct the observed deformations and dynamics.
However, these physics-based priors are mostly derived from experimental models that fall short of capturing real-world complexity.
For example,
in MPM,
the explicit Euler time integration can become unstable in long-term predictions
and the frequent interpolations between particles and grids inevitably lead to an increase of errors;
the mass-spring system is only a rough approximation of true elasticity, limiting their range of realistic applications.

In contrast to existing methods, we treat Gaussian kernels as continuous pieces of matter and use continuum mechanics to model their deformation, making this approach broadly applicable. This formulation allows us to handle real-world scenarios without relying on idealized assumptions, resulting in a more general and robust model.

%% file: sec/3_method.tex
\input{tab_fig/3_overview_v2}

\section{Methodology}

Continuum mechanics is a branch of physics that studies the behavior of materials by modeling them as continuous, rather than discrete, matter. It assumes that materials are continuous and uniform, allowing the analysis of their mechanical behavior (e.g., stress, strain, deformation) under various forces and conditions,
which is generally applicable in real world. 
In this work, instead of discrete particle, we treat each Gaussian kernel as a CMS that describes continuous piece of matter, with a detailed explanation provided in the Appendix.
Hence, our approach can model the deformation and dynamics of objects more accurately. This aligns with the principles of continuum mechanics, which focus on how materials respond to forces and deformations in a continuous domain.
In the following sections, we explain how this is achievable on objects represented as Gaussian Splatting.

\subsection{Problem Formulation}\label{sec:problem}
As depicted in \Figref{fig:overview}(a), given an object reconstructed through Gaussian Splatting, $\mathcal{G} = \{\vx_k, \bm{\sigma}_k, \vc_k, \alpha_k\}_{k\in\mathcal{K}}$
at time $\{t-1,t\}$, where $\vx_k, \bm{\sigma}_k, \vc_k, \alpha_k$ denote the positions, covariances, colors, and opacity, respectively,
and the corresponding attributes $\mathcal{A}=\{\rho_k, \va_k\}_{k\in\mathcal{K}}$,
\mname~aims to predict the Gaussian kernels' states at time $t+1$:
\begin{eqnarray}
    \mathcal{G}_{t+1} &=& \psi_\theta (\mathcal{G}_{t}, \mathcal{G}_{t-1}, \mathcal{A}),\label{eq:iter}
\end{eqnarray}
where $\va_k \in \mathbb{R}^{d}$ is the material attribute vector with $d$ dimensions describing the deformation properties,
and $\rho_k$ is the density for the $k$-th kernel.
The kernels' opacity $\alpha_k$ and attributes $\mathcal{A}$ are invariant over time.
$\psi_\theta$ is our neural network-based \mname~parameterized by $\theta$. More details about \mname's implementation can be found in the appendix.

In each dynamic sequence,
the motion of a deformable object always starts from the template states $\mathcal{G}_{0}$ at time $t=0$,
which is the material space in continuum mechanics.
To approximate $\mathcal{\bar{G}}_{-1}$ and initialize the predictions in \Eqref{eq:iter},
we set the positions $\{\bar{\vx}_{k,-1}\}_{k\in\mathcal{K}}$ as trainable variables,
given which we predict $\{\bm{\bar{\sigma}}_{k,-1}, \bar{\vc}_{k,-1}\}_{k\in\mathcal{K}}$ accordingly.
Additionally,
the unknown attributes $\bar{\mathcal{A}}=\{\bar{\rho}_k, \bar{\va}_{k}\}_{k\in\mathcal{K}}$ are also trainable.
To enable \mname~to learn real-world dynamics,
we render the deformed Gaussian kernels as images for a given view directions $\vd$,
and adopt the multi-view videos $\{I_{t}^{\vd}\}_{t, \vd}$ as supervising signals for training.
More training details can be found in \Secref{sec:training}.

For simplicity,
we omit the timestamp index in the following sections where there is no risk of ambiguity.

\subsection{Integrating with Continuum Mechanics}\label{sec:continuum}
Instead of treating Gaussian kernels as discrete particles for particle-based simulations,
we consider each kernel as a CMS describing continuous piece of matter or a subset of the overall material domain being simulated.
Consequently, the volumes represented by the Gaussian kernels are accounted for in our simulation, aligning naturally with the principles of continuum mechanics.

In continuum mechanics,
the deformed states are defined by the deformation map $\vx=\phi(\mX, t)$,
where $\mX$ are the un-deformed states in material space.
For a Gaussian kernel at $\vx_r$,
we can derive the positions of other kernels nearby at any timestamp through the first-order approximation of the deformation map:
\begin{eqnarray}
    \vx_k &\approx& \vx_r+\mF_k(\mX_k-\mX_r), \label{eq:base_sim}
\end{eqnarray}
where $\mF_k$ is the deformation gradient predicted by our \mname~through the modeling of nearby kernels' interactions.
In practice,
we choose the position of a static Gaussian kernel,
such as the root of a flower that is invariant over time, as $\vx_r$.

Furthermore,
as is favored in the graphics/mechanics,
we represent the deformation gradient in the form of ``Polar SVD'' \cite{DBLP:conf/sca/IrvingTF04, mcadams2011computing, gast2016implicit}:
\begin{eqnarray}
    \mF &=& \mU\bm{\Lambda}\mV^{\top}, \label{eq:svd}
\end{eqnarray}
where $\mU$ and $\mV$ are rotation matrices,
and $\bm{\Lambda}$ is the diagonal matrix for singular values representing the magnitude of the deformation.
In practice,
we predict two quaternions to represent the $\mU$ and $\mV$,
and a 3D vector $\bm{\lambda}=[p, q, r]$ to represent $\bm{\Lambda}$.

Since each kernel represents a tiny piece of matter,
we deform the kernel based on $\mF_k$,
which is demonstrated by the transformations of covariance and color as mentioned in Xie~\etal~\cite{DBLP:conf/cvpr/XieZQLF0J24}:
\begin{eqnarray}
    \bm{\sigma}_k &=& \mF_k\bm{\Sigma}_k\mF_k^{\top},\label{eq:cov}\\
    \vc_k &=& g(\mR_k^{\top}\vd), \qquad \mR_k=\mU_k\mV_k^{\top},\label{eq:color}
\end{eqnarray}
where $\vd$ indicates the view direction, $\mR_k$ is the rotation caused by deformation,
$g$ is the function to evaluate the color represented by spherical harmonic.

Our formulations benefit the simulation of Gaussian kernels as follows.
First,
since the deformed states are computed based on the material space,
\mname~minimizes the impact of error accumulation in long-term predictions,
leading to more robust performance.
Second,
we predict deformation gradients in the form of ``Polar SVD'',
which naturally extend to compute physics variables,
such as the rotation matrix in \Eqref{eq:color},
reducing the need for additional computations to decompose the gradients.
Third,
each component in \Eqref{eq:svd} is interpretable
and can be easily regularized.
For instance,
we represent rotation matrices using quaternions.

\subsection{Hierarchical Structure}\label{sec:hierarchical}
Reconstructed objects often contain densely distributed Gaussian kernels. For example, an object could have 9,000 to 34,000 kernels,
resulting in significant computational overhead when predicting kernel-wise deformation gradients.
Our key insight is that
nearby kernels exhibit similar deformations.
To leverage this,
we build a hierarchical structure by grouping nearby kernels into Center of Mass Systems (CMS).
Each CMS can be regarded as a larger kernel for simulation.
As illustrated in \Figref{fig:overview}(b),
we simulate the dynamics iteratively, starting from the highest level (the coarsest representation), and propagate updates to the physics properties down to the lower levels, reaching individual Gaussian kernels.

\noindent\textbf{Formulation.}
In our approach,
kernels are clustered based on their distances to construct the CMS from the bottom-up:
\begin{eqnarray}
    \vx_{c_{l}} &=& \frac{\sum_{i}(m_{i}\vx_{i})}{m_{c_{l}}},\label{eq:weighted_pos} \qquad \va_{c_{l}}=\frac{1}{N_l}\sum_{i} \va_i,\\
    m_{c_l}&=&\sum_{i}m_{i}, \label{eq:cms_mass} \quad
    \rho_{c_{l}} = \frac{m_{c_{l}}}{V_{c_{l}}}, \label{eq:density} \quad V_{c_{l}}=\sum_{i}V_i, \label{eq:cms_volume}
\end{eqnarray}
where $i\in\mathcal{N}_l$ are kernels to build the hierarchy at $l$-level,
$N_l$ is the number of the neighbors,
$m_{c_{l}}$ and $V_{c_l}$ are mass and volume, respectively.
When $l=0$, $\vx_{c_0}$ refers to the position of the Gaussian kernel,
which can be equivalently regarded as a CMS for continuous piece of matter with volume $V_{c_0}=\sqrt{\det(2\pi\bm{\sigma}_{c_0})}$
and preserves several properties as discussed in \Secref{sec:phys}.
More details can be found in the Appendix.

Consequently,
the constructed CMSs become coarser representations of the objects,
and are treated as simulation units by \mname.
We then apply \Eqref{eq:base_sim} to simulate these units from top-to-bottom and recursively update all CMSs in lower levels following \Eqref{eq:cov} and \Eqref{eq:color}.
By expanding the equations for hierarchy with $L$ levels in total,
we obtain the final form:
\begin{eqnarray}
    \hat{\vx}_{k}^{h-1} &=&  \hat{\vx}^{h}_{c_h}+\prod_{j=h}^{L}\mF^{j}_k \left(\mX_{k}-\mX_{c_{h}}\right),\label{eq:pred_kernel}\\
    \hat{\vx}^{h}_{c_h} &=& \vx_{r}+\sum_{i=h}^{L}\prod_{j=i}^{L}\mF^{j+1}_k \left(\mX_{c_{j}}-\mX_{c_{j+1}}\right),\label{eq:pred_anchor}\\
    \hat{\bm{\sigma}}_{k}^{h-1} &=& \left(\prod_{i=h}^{L}\mF^{i}_k\right)\bm{\Sigma}_k\left(\prod_{i=h}^{L}\mF^{i}_k\right)^{\top},\label{eq:pred_cov}\\
    \hat{\vc}_k^{h-1} &=& g\left(\prod_{i=h}^{L}(\mR_k^i)^{\top}\vd\right), \label{eq:pred_color}
\end{eqnarray}
where the superscript $h-1$ denotes the variables that are updated given $h$-th level's simulation,
while the subscript $c_{h}$ and $k$ indicate the $h$-th level's CMS and Gaussian kernels respectively.
$h\in \{1, \cdots, L\}$ in our study.
When $h=1$,
the outputs $\{\hat{\vx}^0_k, \hat{\bm{\sigma}}_{k}^{0}, \hat{\vc}_k^{0}\}$ represent the deformed states of the Gaussian kernels.
When $h=L$,
$\hat{\vx}^L_{c_L}$ and $\mX_{c_L}$ are both the root kernel $\vx_r$.
The equations are similar in form to those without a hierarchical structure, as described in \Secref{sec:continuum}, with the primary difference being the inclusion of the product of multiple deformation gradients.
Detailed deduction can be found in the Appendix.

\noindent\textbf{Advantages.}
Our hierarchical structure holds several advantages.
First,
the simulation using our hierarchy enables the transition from coarse representations to detailed kernels, while remaining interpretable through explicit equations.
Such formulations integrate seamlessly with Gaussian representations,
clearly capturing the transformations of covariances and colors at each level of the hierarchy, offering good performance with high fidelity.
Second,
the hierarchy offers a meaningful interpretation of the Gaussian kernels,
which can be equivalently regarded as CMSs for continuous pieces of matter,
aligned with the concept of the simulation for continuum mechanics instead of discrete particles.
Third,
the hierarchical structure significantly reduces the redundant predictions for each kernel and enables efficient simulation. 
Suppose that each CMS at level $h$ contains $\gamma_h$ number of CMSs at level $h-1$,
the number of predicted deformation gradients are:
$N_{\mF} = N_{\mathcal{K}} \cdot \sum_{i=1}^{L} (\prod_{j=1}^i\gamma_{j})^{-1}$,
where $N_{\mathcal{K}}$ is the number of Gaussian kernels.
As the Gaussian kernels are densely distributed and share similar deformations with neighbors,
we avoid applying \Eqref{eq:base_sim} at the zero-th level that is equivalent to predicting kernel-wise deformation gradients,
and only simulate the CMSs from the top level down to the first level,
reducing the number of predictions by around 95\%, as shown in our experiments.

\subsection{Explicit Physics Constraints}\label{sec:phys}
\paragraph{Conservation of Mass.}\label{sec:conserv_mass}
Since the density of the object is invariant over time in our study,
the changes of mass are determined by the variations of volumes as shown in \Eqref{eq:density}.
To ensure mass conservation,
the volumes must remain constant during deformations.
In continuum mechanics,
the relationship between the deformed volumes $dv$ and the original volumes $dV$ in material space is expressed as follows
\begin{eqnarray}
    dv &=& \det(\mF)dV.
\end{eqnarray}
Therefore,
by forcing $\det(\mF)=1$,
the volumes remain unchanged,
resulting in mass conservation.
While the rotation matrices in \Eqref{eq:svd} do not affect the determinants of deformation gradients,
we normalize the predicted diagonal matrix $\Lambda$ through the 3D vector $\bm{\lambda}$ as follows:
\begin{eqnarray}
    \bm{\lambda} &=& \left[\frac{p}{\sqrt[3]{pqr}}, \frac{q}{\sqrt[3]{pqr}}, \frac{r}{\sqrt[3]{pqr}}\right], \label{eq:mass_conserve}
\end{eqnarray}
where $p, q, r$ are meaningful only when they are positive numbers.
In the hierarchical structure in \Secref{sec:hierarchical},
since each kernel is mass and volume conservative,
the mass system constructed by \Eqref{eq:weighted_pos} and \Eqref{eq:cms_volume} holds the conservation of mass and volume as well.

\paragraph{Conservation of Momentum.}\label{sec:conserv_mom}
While the CMS is built following \Eqref{eq:weighted_pos},
the deformed CMSs from lower level are expected to hold the momentum of the current CMS as well:
\begin{eqnarray}
    m_{c_{l}}\frac{d\hat{\vx}_{c_{l}}}{dt} &=& \sum_{i\in \mathcal{S}_{c_l}}m_{i}\frac{d\hat{\vx}_{i}}{dt},
\end{eqnarray}
where $i\in \mathcal{S}_{c_l}$ denote the CMSs that are from level $l-1$ and belong to the $c_l$-th CMS.
We thus apply the square error to constrain the relations between the hierarchical layers equivalently as follows:
\begin{eqnarray}
    \mathcal{L}_{\mathrm{mom}} &=& \sum_{l=1}^{L-1}\lVert m_{c_l}\hat{\vx}_{c_l} - \sum_{i\in\{c_{l-1}\}}(m_{i}\hat{\vx}_{i})\rVert_2^2. \label{eq:mom_loss}
\end{eqnarray}

\subsection{Training}\label{sec:training}
To capture the underlying physics laws from reality,
we adopt multi-view videos $\{I_{t}^{\vd}\}_{t, \vd}$ to supervise the training of \mname:
\begin{align}
    \hat{I}_{t}^{\vd} &= \mathcal{F}(\hat{G}_{t}, \vd),\\
    \mathcal{L}_{I,t} &= \lambda \mathcal{L}_2(\hat{I}_{t}^{\vd}, I_{t}^{\vd}) + (1-\lambda)\mathcal{L}_{\mathrm{D-SSIM}}(\hat{I}_{t}^{\vd}, I_{t}^{\vd}),
\end{align}
where the predicted Gaussian kernels $\hat{G}_{t}$ are rendered by function $\mathcal{F}$ for the given view $\vd$,
$\mathcal{L}_2$ represents the $\ell$2 errors between images,
and $\lambda$ is a hyper-parameter to balance the loss weight.
During training,
\mname~auto-regressively predicts $T$ steps,
where $T$ starts with $T=1$ and increases as training progresses.
The gradients are back-propagated between two adjacent time steps to avoid gradient explosion.

Dynamic predictions at the start of the training with randomly initialized neural networks can be challenging for \mname.
To this end, we introduce a static loss to assist the initialization in a self-supervised manner:
\begin{eqnarray}
    \hat{\mathcal{G}}_0 &=& \psi_\theta(\mathcal{G}_0, \mathcal{G}_0),\\
    \mathcal{L}_{\mathrm{static}} &=& \sum_k \lVert \hat{\vx}_{k, 0} - \vx_{k, 0}\rVert^2_2, \label{eq:static}
\end{eqnarray}
which suggests that given static inputs $\mathcal{G}_0$ with zero velocities,
the predicted states of Gaussian kernels should remain unchanged.
We apply \Eqref{eq:static} with a probability of $1/T$ to balance the static predictions and dynamic signals. 
The final training loss is summarized as follows:
\begin{eqnarray}
    \mathcal{L} &=& \sum_{t=1}^{T}\mathcal{L}_{I,t}+\mathcal{L}_{\mathrm{static}}+\mathcal{L}_{\mathrm{mom}}. \label{eq:loss}
\end{eqnarray}

%% file: tab_fig/3_overview_v2.tex
\begin{figure*}[t]
        \begin{center}
            \includegraphics[width=0.92\textwidth]{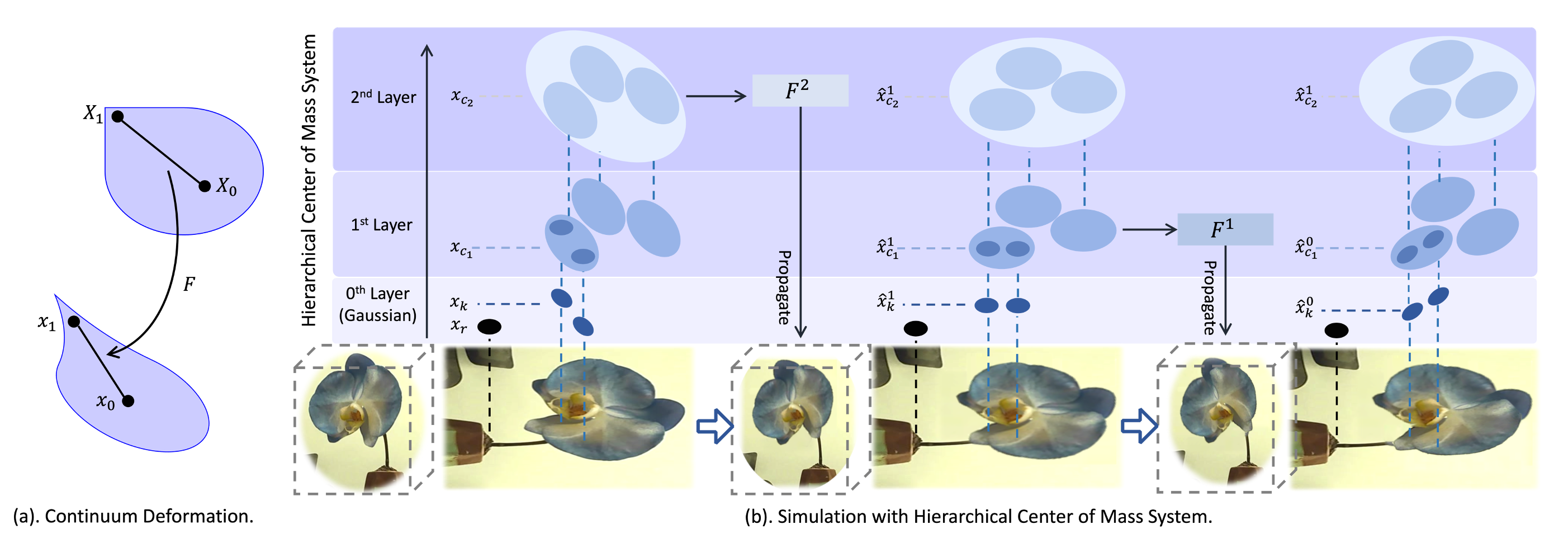}
        \end{center}
        \vspace{-8mm}
        \caption{\small{
            (a). \mname~is designed based on the continuum mechanics.
            The deformations of the continuum are described through deformation gradients $\mF$.
            (b). Example with $L=2$ levels of hierarchical structure, which is built by constructing the Center of Mass Systems from the bottom up.
            \mname~iteratively simulates dynamics from the top level, predicting the corresponding deformation gradients $\mF$,
            which are then propagated down to the lower levels till individual kernels to update their
            dynamics.
        }}
        \vspace{-6mm}
        \label{fig:overview}
\end{figure*}

%% file: sec/5_experiments.tex
\section{Experiments}
\input{tab_fig/4_quant_compare}
\input{tab_fig/4_dynamic_cmp}

\noindent\textbf{Real Dataset.}
As shown in \Figref{fig:teaser},
we collect four real objects that typically preserve elasticity and are common in daily life, namely,
moth-orchids, carnation, duck, and pudding.
The objects are randomly poked and dragged through various directions with different external forces,
resulting in 30 dynamic sequences of 50 FPS for each object.
Each sequence consists of 4 to 6 views captured by different cameras,
which are synchronized through time code during video recording.
To assist the reconstruction of each object,
we further gather 200 images of the static scenes.
Camera poses are obtained through the off-the-shelf Structure-from-Motion toolkit COLMAP \cite{DBLP:conf/cvpr/SchonbergerF16, DBLP:conf/eccv/SchonbergerZFP16}.
Since we focus on the temporal dynamics of the objects,
we segment out the foreground using SAM \cite{DBLP:journals/corr/abs-2408-00714}
to improve the training efficiency.
For each object,
we regard the first $T=16$ frames of all sequences as the training data,
while taking the remaining 34 or 83 frames for the test.

\noindent\textbf{Synthetic Dataset.}
We synthesize elastic dynamics for the bunny using Blender\footnote{\url{https://www.blender.org/}},
and adopt \dname's settings to render multi-view videos
except that this domain is 24 FPS.
The synthetic dataset includes 30 sequences with 1800 dynamic frames in total.

\noindent\textbf{Baselines.}
We adopt DreamGaussian4D (DG4D) \cite{DBLP:journals/corr/abs-2312-17142} and PhysDreamer (PD) \cite{DBLP:journals/corr/abs-2404-13026} as our baseline.
PhysGaussian \cite{DBLP:conf/cvpr/XieZQLF0J24} requires experts-defined material parameters to generate reasonable deformations.
Since PhysGaussian and PD both adopt the same kind of MPM simulation engine,
PhysGaussian can generate similar results given the physics parameters estimated by PD.
Thus PD can represent PhysGaussian's performance.
For PD, we use the first three frames to optimize the velocity and 16 frames to estimate the material parameters.
In addition,
since DG4D cannot predicts dynamics beyond training set,
we loop the results for comparisons of long-term predictions.
We ensure that all models are trained on the same amount of frames.

\noindent\textbf{Evaluation Metric.}
Since our dataset is either collected from the real world or simulated through physics,
it is unnecessary to evaluate the motion realism and video quality as in video diffusion models \cite{DBLP:journals/corr/abs-2311-15127}.
Instead,
we can directly measure the pixel-wise errors,
which is the $\ell$2 norm in our study,
to validate the faithfulness of the predictions.
We average the $\ell$2 errors on each sequence
and obtain the mean of errors from all sequences as the final result.
We further report the SSIM for structural similarities and LPIPS for perceptual evaluations in the supplementary material.
All evaluation processes are conducted on objects only,
where the background in ground truth images is masked and omitted.
We combine the foreground and background together and adjust the brightness for better visualization.

\input{tab_fig/4_qual_compare}

\subsection{Dynamic Simulations}
While our \mname~naturally supports interactive simulations as illustrated in supplementary material, we focus on the ability of mimicking and foreseeing motions in reality.
We report the quantitative and qualitative results in \Tabref{tbl:quant_cmp} and \Figref{fig:qualitative}, respectively.
Videos are provided in the supplementary for better comparisons.

\noindent\textbf{Dataset's Dynamic Patterns and Model Behaviors.}
As shown in \Figref{fig:sp_time_slice},
due to the dissipation in reality,
objects in \dname~tend to stop moving as time goes by.
On ``Bunny'' domain,
we reduce the damping effects to generate elastic dynamics lasting longer.
While models with faithful predictions achieve lower errors,
there are short-cuts to achieve abnormally lower $\ell$2 error:
motions with fast dissipation or even static predictions,
which will generally lead to larger overlapping with the ground truth.
Smaller dissipation of the ground truth mitigates the short-cut impact and leads to slower decrease of errors.
Therefore,
effective evaluations must combine both qualitative and quantitative results.
Moreover,
since the contrast between the foreground color and black background is larger in ``Bunny'',
the variations of absolute errors are more obvious than other domains. Please refer to supplementary for more details.

As shown in \Tabref{tbl:quant_cmp},
we divide the evaluations into two parts:
1. ``Seen'' represents the frames during training from $t=0$ to $t=15$, indicating the quality of estimating the initial velocities and the training loss;
2. ``Unseen 34/83'' refers to the 34/83 frames of unseen motions from $t=16$ to $t=49$ and $t=16$ to $t=99$ respectively,
suggesting the generalization abilities on unknown dynamics.

As shown in \Figref{fig:qualitative},
while DG4D can reconstruct the dynamics in the training set and obtain reasonable errors in \Tabref{tbl:quant_cmp}, it cannot predict dynamics beyond what has been captured during training. In particular, the method fails to predict the dissipation of motions and yields higher errors on test frames.
Another baseline, PD, can physically constrain motions,
such as the swaying of ``Carnation''. However, it struggles to replicate the complex deformations, such as those on ``Mothorchids'', and produces motions with rapid dissipation across all domains.
This results in large errors on the training set and an abnormally fast decline in errors on the test set.
In contrast,
our \mname~faithfully predict the dynamics and achieves lower errors in all cases,
suggesting the effectiveness of our method.

We further demonstrate the dynamic details using space-time slices as shown in \Figref{fig:sp_time_slice},
where the vertical axis denotes time and the horizontal axis represents a spatial slice of the object.
Neither DG4D nor PD produce satisfactory results - DG4D fails to predict unseen dynamics, while PD exhibits rapid dissipation, especially for smaller objects with higher swaying frequencies.
On the contrary, \mname~effectively and robustly simulates elastic deformations across various types.

\input{tab_fig/4_efficiency}
\noindent\textbf{Computational Efficiency.}
As shown in \Tabref{tbl:speed} where we use ``Mothorchids'' for demonstration,
our hierarchical structure enables us to reduce the kernel-wise computations by 95\%,
and achieve both faster prediction speed and lower memory cost on one NVIDIA A100 GPU.

\subsection{Ablation Study}\label{sec:abl}
We investigate the effectiveness of our physically interpretable designs,
including the mass and momentum conservation in \Eqref{eq:mass_conserve} and \Eqref{eq:mom_loss}.
Models are trained and tested on ``Mothorchids'' domain,
which is challenging and involves sophisticated deformations.
Moreover,
we further verify the generalization abilities of \mname~by jointly training our model on all domains,
which is denoted by ``Jointly''.

As shown in \Tabref{tbl:model_abl} and \Figref{fig:abl_qual},
\mname~trained on all domains delivers similar accuracy both quantitatively and qualitatively compared with \mname~trained solely on ``Mothorchids'',
suggesting that \mname~is highly generalizable on different objects thanks to our shape-independent designs.
Since ``Duck'' introduces more static motions that affect the balance of dynamics data,
``Jointly'' tends to predict motions with slightly faster dissipation,
leading to a decrease in errors on the test set.
\mname~without constraint of mass conservation displays higher errors
and struggles with maintaining the shape or volume of the flower as shown in \Figref{fig:abl_qual}.
Without the momentum constraint in \Eqref{eq:mom_loss},
\mname~yields blurred details with less dynamic motions.
In contrast,
\mname~with both physics constraints obtains vivid and robust performance,
suggesting the effectiveness of our explicit physics constraints.

%% file: tab_fig/4_quant_compare.tex
\begin{table*}[ht]
\setlength{\tabcolsep}{3pt}
\caption{
\small{
$\ell$2 errors ($1\times10^{-3}$) against the ground truth videos on our dataset \dname,
which consists of ``Mothorchids'', ``Carnation'', ``Pudding'', and ``Duck'',
as well as synthetic ``Bunny''.
$\ell$2 errors on training frames from $t=0$ to $t=15$ is denoted by ``Seen''.
For the unseen frames,
we report the errors on 83 and 34 frames from $t=16$ to $t=99$ and $t=16$ to $t=49$ respectively.
Since DG4D is unable to predict unseen dynamics,
we loop the results from training set as reference.
PD struggles with the challenging deformations and delivers dynamics with fast dissipation, leading to higher errors on training set and abnormally lower errors on unseen frames.
Our \mname~achieves superior performance in all cases.
}}
\vspace{-6mm}
\label{tbl:quant_cmp}
\begin{center} \footnotesize 
\begin{tabular}{lcccccccccc}
\toprule
\multirow{2}{*}{\bf Methods}	&\multicolumn{2}{c}{\bf{Mothorchids}}    & \multicolumn{2}{c}{\bf Carnation} & \multicolumn{2}{c}{\bf Pudding} & \multicolumn{2}{c}{\bf Duck} & \multicolumn{2}{c}{\bf Bunny} \\ \cmidrule(lr{.75em}){2-3}\cmidrule(lr{.75em}){4-5}\cmidrule(lr{.75em}){6-7}\cmidrule(lr{.75em}){8-9}\cmidrule(lr{.75em}){10-11}
						&Seen		&Unseen 83 		&Seen		         &Unseen 83		&Seen 		&Unseen 83          &Seen		&Unseen 34 & Seen & Unseen 34
\\ \cmidrule(lr{.75em}){1-1}\cmidrule(lr{.75em}){2-3}\cmidrule(lr{.75em}){4-5}\cmidrule(lr{.75em}){6-7}\cmidrule(lr{.75em}){8-9}\cmidrule(lr{.75em}){10-11}
\bf DG4D \cite{DBLP:journals/corr/abs-2312-17142}		    & 3.26$\pm$0.61		& 11.32$\pm$3.09	        & 6.20$\pm$0.85    &31.63$\pm$5.10    &2.13$\pm$0.23		&6.71$\pm$0.49   &2.65$\pm$0.59  & 4.61$\pm$1.07 & 1.21$\pm$0.11 & 3.10$\pm$0.63\\
\bf PD \cite{DBLP:journals/corr/abs-2404-13026}		    & 5.97$\pm$0.76		& 2.59$\pm$0.27	        & 7.23$\pm$0.91    &6.15$\pm$0.84    &2.31$\pm$0.34		&1.79$\pm$0.28   &2.73$\pm$0.67  & 2.49$\pm$0.62    &4.77$\pm$0.45  &3.62$\pm$0.31\\\midrule
\bf \mname(Ours) 		& \textbf{1.78$\pm$0.23}		& \textbf{1.85$\pm$0.14}	        & \textbf{3.69$\pm$0.36}  &\textbf{6.02$\pm$0.75}	&\textbf{1.12$\pm$0.01}  &\textbf{1.16$\pm$0.01}		        &\textbf{2.21$\pm$0.48}	&\textbf{2.36$\pm$0.48} & \textbf{1.15$\pm$0.12}  & \textbf{2.47$\pm$0.40}\\
\bottomrule
\end{tabular}
\end{center}
\vspace{-8mm}
\end{table*}

%% file: tab_fig/4_dynamic_cmp.tex
\begin{figure}[t]
        \begin{center}
            \includegraphics[width=0.49\textwidth]{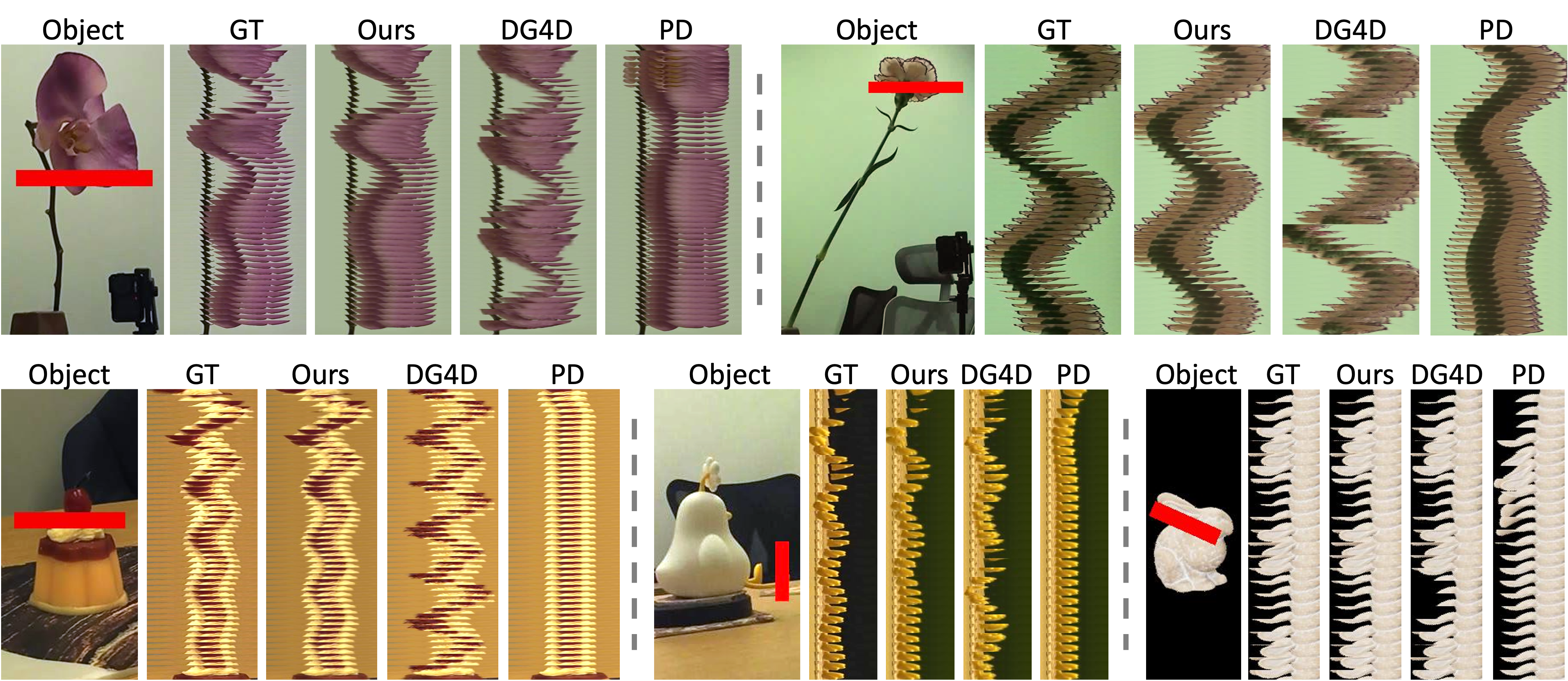}
        \end{center}
        \vspace{-6mm}
        \caption{\small{
        Space-time slices of sampled dynamics.
        Since DG4D cannot predict unseen dynamics and we loop the training results for reference,
        its predictions display fixed dynamic patterns.
        PD exhibits obvious dissipation especially for small objects with high swaying frequency. \mname(Ours) faithfully mimics the deformation patterns in all cases.
        }
        }
        \vspace{-6mm}
        \label{fig:sp_time_slice}
\end{figure}

%% file: tab_fig/4_qual_compare.tex
\begin{figure*}[!h]
        \begin{center}
            \includegraphics[width=0.98\textwidth]{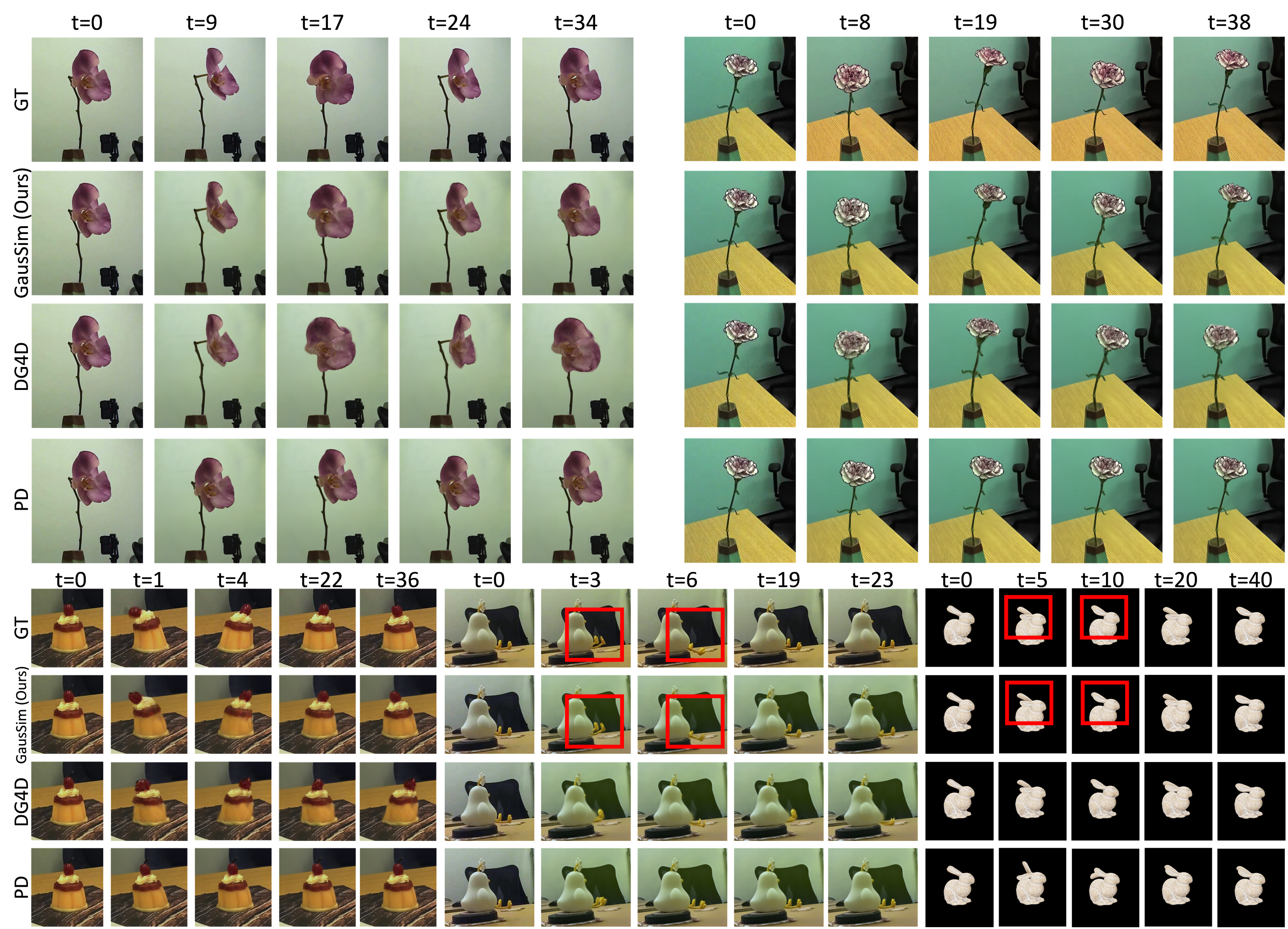}
        \end{center}
        \vspace{-6mm}
        \caption{\small{
        Qualitative comparisons.
        DG4D can only reconstruct the dynamics within the training set, while PD faces difficulties in mimicking the challenging deformations and generates motions with fast dissipation. \mname~achieves faithful and robust predictions regardless of the complexity of the dataset.
        Note that the images in the ``Duck'' domain differ slightly from the background of the ground truth. This discrepancy stems from the quality of the 
        Gaussian reconstruction, which is unrelated to our primary focus on
        simulations.
        }
        }
        \vspace{-5mm}
        \label{fig:qualitative}
\end{figure*}

%% file: tab_fig/4_efficiency.tex
\begin{table}[t]
\caption{
\small{
Efficiency test on ``Mothorchids''.
Left: Number of Gaussian kernels/CMS at each hierarchy.
We compute the ratio between total number of predictions $N_{\mF}$
in \Secref{sec:hierarchical}
and the number of Gaussian kernels $N_{\mathcal{K}}$ as $N_{\mF}/N_{\mathcal{K}}$.
Our hierarchical structure reduces the kernel-wise computations by around 95\%.
Right: Per-frame forward time and GPU memory cost averaged on 50 frames of predictions.
\mname~without hierarchy is marked by ``w/o H''.
}}
\vspace{-2mm}
\begin{subtable}[t]{0.16\textwidth}
    \setlength{\tabcolsep}{1.8pt}
    \centering
    \footnotesize
    \begin{tabular}{lc}
    \toprule
    \bf Level	    & \bf Amount
    \\ \midrule
    l=0  & 23422\\
    l=1	& 1203\\
    l=2	& 11\\\midrule
    $N_F/N_{\mathcal{K}}$	& 0.05\\
    \bottomrule
    \end{tabular}
\end{subtable}
\hfill
\begin{subtable}[t]{0.32\textwidth}
    \setlength{\tabcolsep}{1.8pt}
    \centering
    \footnotesize
    \begin{tabular}{lcc}
    \toprule
        & \bf Time (s)    & \bf GPU (GB)
    \\ \midrule
    DG4D \cite{DBLP:journals/corr/abs-2312-17142} & 0.14$\pm$0.01 & 7.6\\
    PD \cite{DBLP:journals/corr/abs-2404-13026}	& 1.67$\pm$0.05  &4.6\\\midrule
    \mname~w/o H	& 0.36$\pm$0.01    &3.5\\
    \mname	& \textbf{0.13$\pm$0.01}  &\textbf{2.1}\\
    \bottomrule
    \end{tabular}
\end{subtable}
\vspace{-3mm}
\label{tbl:speed}
\end{table}

%% file: sec/6_conclusion.tex
\section{Discussion}
\input{tab_fig/4_abl_quant}
\input{tab_fig/4_abl_qual}

In this paper, we aimed to learn the underlying physical laws of real-world objects represented through Gaussian Splatting \cite{DBLP:journals/tog/KerblKLD23} using multi-view videos. By treating Gaussian kernels as Center of Mass Systems (CMS) that govern continuous pieces of matter, we integrated continuum mechanics with neural networks and introduced \mname~to capture dynamic deformations accurately. To achieve efficient, high-fidelity simulations, we employed a hierarchical structure based on CMS, enabling a coarse-to-fine simulation approach. We further constrained \mname~with explicit mass and momentum conservation principles, ensuring robust and physically plausible dynamics.
In addition, we presented a new dataset, \dname, featuring real-world objects like ``Mothorchids'', ``Carnation'', ``Pudding'', and ``Duck''. We also provide synthetic ``Bunny'' using Blender. Experiments show that \mname~faithfully reproduces dynamics that adhere closely to the ground truth, effectively capturing the real-world physical laws. 
Our method is not free of limitations. The training process requires high-quality data; for instance, capturing high-speed motions necessitates increasing the camera's shutter speed to reduce image blur. Additionally, accurate learning of object dynamics requires minimizing background noise in the segmented foreground. Furthermore, since \mname~is based on Gaussian kernels,
the performance can be affected by the quality of the Gaussian Splatting reconstruction, particularly impacting the object's surface color.
Despite these limitations,
our results demonstrate that \mname~offers a promising approach for realistic and interpretable simulations.

\noindent
\textbf{Acknowledgement.}
This study is supported under the RIE2020 Industry Alignment Fund  Industry Collaboration Projects (IAF-ICP) Funding Initiative, as well as cash and in-kind contributions from the industry partner(s). It is also supported by Singapore MOE AcRF Tier 2 (MOE-T2EP20221-0011),
and partially funded by the Shanghai Artificial Intelligence Laboratory and The University of Hong Kong Startup Fund.

%% file: tab_fig/4_abl_quant.tex
\begin{table}[t]
    \caption{
    \small{
    Ablation studies in terms of $\ell$2 errors ($1\times10^{-3}$) on ``Mothorchids'' domain,
    which is challenging with more complex deformations.
    We investigate the effectiveness of the explicit physics constraints,
    namely the mass conservation in \Eqref{eq:mass_conserve} and the momentum conservation in \Eqref{eq:mom_loss},
    which are denoted by ``w/o mass'' and ``w/o momen'', respectively.
    Furthermore,
    we jointly train our \mname~on all domains to illustrate the generalization abilities,
    which is denoted by ``Jointly''.
    Quantitatively,
    \mname s trained on either ``Mothorchids'' solely or all domains jointly deliver similar $\ell$2 errors,
    suggesting the generalization abilities of our method on diverse objects.
    }}
    \vspace{-6mm}
    \label{tbl:model_abl}
    \setlength{\tabcolsep}{1.8pt}
    \begin{center}
    \small
    \begin{tabular}{lcccc}
    \toprule
    \bf \mname  & \bf Full Model 	    & \bf Jointly	& \bf w/o mass   & \bf w/o momen
    \\\cmidrule(lr{.75em}){1-1}\cmidrule(lr{.75em}){2-2}\cmidrule(lr{.75em}){3-5}
    Seen  & \textbf{1.78$\pm$0.23}   & 1.90$\pm$0.54 & 2.19$\pm$0.52 &2.64$\pm$0.62\\
    Unseen 83	&\textbf{1.85$\pm$0.14} 	& 1.88$\pm$0.18    &1.90$\pm$0.16 &1.91+0.17\\
    \bottomrule
    \end{tabular}
    \end{center}
    \vspace{-6mm}
\end{table}

%% file: tab_fig/4_abl_qual.tex
\begin{figure}[t]
    \vspace{-1mm}
        \begin{center}
            \includegraphics[width=0.45\textwidth]{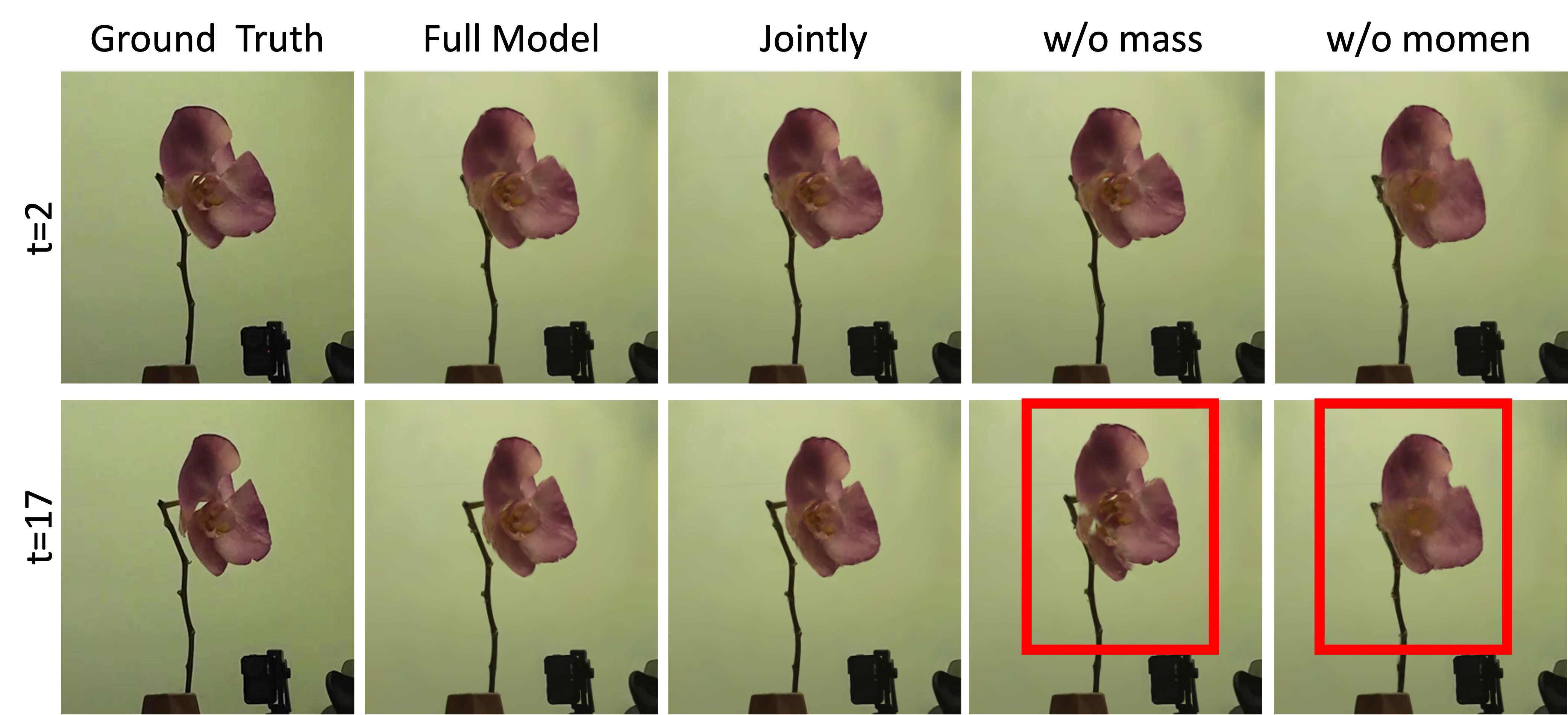}
        \end{center}
        \vspace{-6mm}
        \caption{\small{
        Visualizations of the results in the ablation study.
        \mname~jointly trained on all domains is denoted by ``Jointly''.
        \mname~trained solely on ``Mothorchids'' and jointly on all domains both obtain similar dynamic details comparing with the ground truth.
        \mname~without mass conservation struggles with maintaining the shape or volume of the flower,
        while \mname~without momentum constraint generates blurred surfaces and more static motions starting from frame $t=2$.
        With both the physics constraints,
        \mname~can produce realistic deformations compared with ground truth.
        }
        }
        \vspace{-6mm}
        \label{fig:abl_qual}
\end{figure}

%% file: sec/X_suppl.tex
\clearpage
\setcounter{page}{1}
\setcounter{section}{0}
\renewcommand{\thesection}{A\arabic{section}}
\setcounter{table}{0}
\renewcommand{\thetable}{A\arabic{table}}
\setcounter{figure}{0}
\renewcommand{\thefigure}{A\arabic{figure}}

   \newpage
       \twocolumn[
        \centering
        \Large
        \textbf{\thetitle}\\
        \vspace{0.5em}Supplementary Material \\
        \vspace{1.0em}
        \input{tab_fig/app_symbols}
       ] 

\section{Symbols}
As shown in \Tabref{tbl:app_symbols},
we list the symbols used in the main manuscript and supplementary materials.

\section{Methodology Proofs and Details}
In this paper,
we treat each Gaussian kernel as continuous pieces of matter and formulate \mname~based on continuum mechanics.
The hierarchical structure is built following the rules of Center of Mass Systems (CMS).
Most importantly,
the Gaussian kernels themselves are already Center of Mass Systems as shown in \Secref{sec:app_gs_property},
which are constructed for the areas of continuous volume.

\label{sec:app_proof}
\subsection{Proof of Hierarchical Structure}
Suppose we already have the simulated results $\hat{\vx}_k^{h+1}$ at level $h+1$.
By regarding the results $\hat{\vx}_k^{h+1}$ as a special kind of ``template states'' mentioned in \Eqref{eq:base_sim},
the formulations of simulating the lower levels in a recursive manner are as follows:
\begin{eqnarray}
    \hat{\vx}_k^{h} &=& \hat{\vx}_{c_{h+1}}^{h+1} + F^{h+1}_k(\hat{\vx}_k^{h+1}-\hat{\vx}_{c_{h+1}}^{h+1}),\label{eq:app_h}\\
    \hat{\vx}_k^{h-1} &=& \hat{\vx}_{c_{h}}^h + F^h_k(\hat{\vx}_k^h-\hat{\vx}_{c_h}^h),\label{eq:app_hn1}
\end{eqnarray}
where $\hat{\vx}_{c_{h+1}}^{h+1}, \hat{\vx}_{c_h}^h$ are the barycenters of the Center of Mass Systems at level $h+1, h$ respectively.

By expanding the variables in \Eqref{eq:app_hn1} using \Eqref{eq:app_h},
we have:
\begin{eqnarray}
    \hat{\vx}_k^{h-1} &=& \hat{\vx}_{c_{h}}^{h} + F^{h}_k F^{h+1}_k(\hat{\vx}_k^{h+1}-\hat{\vx}_{c_{h}}^{h+1}).\label{eq:app_hn2}
\end{eqnarray}
Therefore,
through the expansion of the variables within the brackets,
the results $\hat{\vx}_k^{h-1}$ at level $h-1$ can be traced back till the top level $L$ as follows:
\begin{eqnarray}
    \hat{\vx}_k^{h-1} &=& \hat{\vx}_{c_{h}}^{h} + \prod_{j=h}^L F^{j}_k(\hat{\vx}_k^{L}-\hat{\vx}_{c_{h}}^{L}),\label{eq:app_L}
\end{eqnarray}
where we define $\hat{\vx}_k^{L}, \hat{\vx}_{c_{h}}^{L}$ as $\mX_k, \mX_{c_h}$, which are from the original material space as the initial conditions for the recursive formulations, respectively.
Specifically,
$\mX_{c_L}$ is defined as the static position $\vx_r$.
Further replacing $\hat{\vx}_k^{L}, \hat{\vx}_{c_{h}}^{L}$ by $\mX_k, \mX_{c_h}$,
we have:
\begin{eqnarray}
    \hat{\vx}_k^{h-1} &=& \hat{\vx}_{c_{h}}^{h} + \prod_{j=h}^L F^{j}_k(\mX_k-\mX_{c_{h}}), \label{eq:app_hsim}
\end{eqnarray}
which is exactly the \Eqref{eq:pred_kernel} in the main manuscript.

\Eqref{eq:pred_anchor} can be obtained in the same manner,
where the only difference is the subscript.
For illustration,
we show an example of expanding the equation by two steps from level $h$ to level $h+2$ as follows:
\begin{eqnarray}
    \hat{\vx}_{c_{h}}^{h} &=& \hat{\vx}_{c_{h+1}}^{h+1} + F^{h+1}_k(\hat{\vx}_{c_{h}}^{h+1}-\hat{\vx}_{c_{h+1}}^{h+1}),\\
    &=& \hat{\vx}_{c_{h+2}}^{h+2}+F^{h+2}_k(\hat{\vx}_{c_{h+1}}^{h+2}-\hat{\vx}_{c_{h+2}}^{h+2}) \\
    & & + F^{h+1}_k F^{h+2}_k(\hat{\vx}_{c_{h}}^{h+2}-\hat{\vx}_{c_{h+1}}^{h+2}) \\
    &=& \vx_{r}+\sum_{i=h}^{L}\prod_{j=i}^{L}F^{j+1}_k \left(\mX_{c_{j}}-\mX_{c_{j+1}}\right).
\end{eqnarray}

As for the covariance in \Eqref{eq:pred_cov} in the main manuscript,
we start from the Gaussian kernel at material space:
\begin{eqnarray}
    G_k(\mX) &=& e^{-\frac{1}{2}(\mX-\mX_k)^\top \bm{\Sigma}_k^{-1} (\mX-\mX_k)} \label{eq:app_gs}
\end{eqnarray}
According to \Eqref{eq:app_hsim},
we have
\begin{eqnarray}
    \mX-\mX_{c_h} &=& \left(\prod_{j=h}^L F^{j}_k\right)^{-1}(\hat{\vx}^{h-1}-\hat{\vx}_{c_h}^h),\\
    \mX_{k}-\mX_{c_h} &=& \left(\prod_{j=h}^L F^{j}_k\right)^{-1}(\hat{\vx}_k^{h-1}-\hat{\vx}_{c_h}^h),\\
    \mX-\mX_{k} &=& (\mX-\mX_{c_h}) - (\mX_{k}-\mX_{c_h}) \\
                &=& \left(\prod_{j=h}^L F^{j}_k\right)^{-1}(\hat{\vx}^{h-1}-\hat{\vx}_k^{h-1}), \label{eq:app_covproof}
\end{eqnarray}
where $\mX$ and $\mX_k$ share the same deformation sequences since $\mX$ belongs to the $k$-th kernel and they are always deformed together.
Therefore,
combining \Eqref{eq:app_covproof} with \Eqref{eq:app_gs},
the Gaussian kernel can be represented by
\begin{eqnarray}
    G_k(\mX) &=& e^{-\frac{1}{2}(\hat{\vx}^{h-1}-\hat{\vx}_k^{h-1})^\top (\hat{\bm{\sigma}}_k^{h-1})^{-1} (\hat{\vx}^{h-1}-\hat{\vx}_k^{h-1})},\\
    \hat{\bm{\sigma}}_k^{h-1} &=& \left(\prod_{j=h}^L F^{j}_k\right)\bm{\Sigma}_k\left(\prod_{j=h}^L F^{j}_k\right)^\top,
\end{eqnarray}
which is the \Eqref{eq:pred_cov} in the main manuscript.

As for the color in \Eqref{eq:pred_color} from the main paper,
we provide an analysis below.
As mentioned above,
for the given simulated results $\hat{\vx}_k^{h+1}$ at level $h+1$,
we treat them as a special kind of "template states",
based on which we predict the new deformation gradients to deform the kernels or Center of Mass Systems in the lower levels.
Thus,
we can always apply \Eqref{eq:color} recursively to compute the new color after each level's simulation,
leading to the form in \Eqref{eq:pred_color}.

\subsection{Gaussian Kernel Is CMS}\label{sec:app_gs_property}
\noindent\textbf{Equivalent Volume.}
Suppose we have a kernel
\begin{eqnarray}
    G(\vx)&=&e^{\frac{1}{2}(\vx-\bm{\mu})^\top \Sigma^{-1}(\vx-\bm{\mu})},
\end{eqnarray}
where we assume that for a given position $\vx$,
$G(\vx)$ is proportional to the density with ratio $\rho$,
which is invariant overtime.
Thus,
the total mass of this kernel,
which is the integration of the Gaussian function,
is known as:
\begin{eqnarray}
    m&=&\int \rho G(\vx) d^3x \\
    &=& \rho\sqrt{\det(2\pi\Sigma)},
\end{eqnarray}
where we obtain the volume for the kernel equivalently as $V=m/\rho=\sqrt{ \det(2\pi\Sigma)}$,
or the volume is proportional to the root of determinant of covariance matrix as $V\varpropto \sqrt{\det(\Sigma)}$ since the trainable $\rho$ can learn the coefficient.

\noindent\textbf{Gaussian kernel is a Center of Mass System for continuous area.}
The barycenter for the Gaussian kernel locates at the mean position $\bm{\mu}$.
To illustrate,
we have the barycenter's position on x-axis,
which should be $\bm{\mu}_x$,
as follows:
\begin{eqnarray}
    && \frac{1}{m}\int x dm\\     &=&\int \frac{x}{m}\cdot\rho G(\vx) dxdydz\\
    &=& \int \frac{x-\bm{\mu}_x}{m}\rho G(\vx) d(x-\bm{\mu}_x) d(y-\bm{\mu}_y)d(z-\bm{\mu}_z) \label{eq:app_firstpart} \\
    && + \int \frac{\bm{\mu}_x}{m}\rho G(\vx) dxdydz\\
    &=&0+\bm{\mu}_x,
\end{eqnarray}
where \Eqref{eq:app_firstpart} equals to $0$ since it is an odd function with symmetric integral domain.
The same applies to $\bm{\mu}_y, \bm{\mu}_z$.

\input{tab_fig/app_radii}
\section{Input Details of \mname}
We adopt the MeshGraphNet \cite{DBLP:conf/icml/Sanchez-Gonzalez20} as our backbone and model the interactions of Center of Mass Systems to predict the corresponding gradient deformations.
The inputs of \mname~includes node features,
which denote the states of the Center of Mass Systems,
and edge features,
which represent the interaction information between neighbor nodes.
Specifically,
the node feature includes:
\begin{itemize}
    \item The equivalent acceleration of the local system, \ie Center of Mass System: $-\frac{\vx_{c,t}-2*\vx_{c,t-1}+\vx_{c,t-2}}{dt^2}$.
    \item The local velocity of each component $k$ within the Center of Mass System: $\frac{(\vx_{k,t}-\vx_{c_t})-(\vx_{k,t-1}-\vx_{c,t-1})}{dt}$.
    \item The learnable attribute for each component $k$: $\va_k$.
\end{itemize}
We assign an edge feature between nodes when they are close to each other in both material space and deformed states.
And the edge feature with source node $\vx_s$ and destination node $\vx_d$ includes:
\begin{itemize}
    \item The ratio of the deformation: $\lVert\vx_{d,t}-\vx_{s,t}\rVert/\lVert\mX_d-\mX_s\rVert$.
    \item The direction from source node to target node: $(\vx_{d,t}-\vx_{s,t})/\lVert\vx_{d,t}-\vx_{s,t}\rVert$.
    \item The relative velocity: $\frac{(\vx_{d,t}-\vx_{s,t})-(\vx_{d,t-1}-\vx_{s,t-1})}{dt}$.
    \item The relative angles given the barycenter $c$: $<\vx_{d,t}-\vx_{c,t},\vx_{s,t}-\vx_{c,t}>-<\mX_d-\mX_c, \mX_s-\mX_c>$.
\end{itemize}

The radii to cluster kernels and construct the Center of Mass Systems are as shown in \Tabref{tbl:app_radii}.

\section{Experiment Details}
\input{tab_fig/app_color}
\subsection{Dataset \dname}\label{sec:app_dataset}
To capture the dynamic motions with less blurred details,
we increase the camera's shutter speed,
which inevitably results in darker images.
Since the brightness does not affect our study,
we adopt a post-processing step to increase the brightness for all images shown in the paper and supplementary for better visual quality.
In addition,
we segment out the foreground and focus on the motions of the objects.
Training and evaluations are conducted using the segmented images without post-processing.
As shown in \Tabref{tbl:app_contrast},
we report the averaged color values normalized between 0 to 1 for the foreground objects.
Notice that the black background's color value is $(0,0,0)$,
and the contrast between the foreground color and black background is larger in ``Bunny''.
This phenomenon results in larger variations of absolute $\ell$2 errors in the quantitative comparisons.

\noindent\textbf{Differences from Video Diffusion-based Data.}
Instead of distilling priors from Video Diffusion Model,
our data is completely from real world,
without concerning about the realism of the videos generated by diffusion models.
\mname~trained on our dataset can directly replicate and foresee the dynamics in real world,
closing the gap between experimental settings and real-world scenarios.

\subsection{Implementation Details}
\noindent\textbf{\mname.}
We adopt the MeshGraphNet \cite{DBLP:conf/icml/Sanchez-Gonzalez20} consisting of 16 graph neural layers as \mname's backbone to handle the kernel-wise interactions.
The output deformation gradients are obtained through a three-layer MLP,
with the output dimension being set to 11,
where 8 for two quaternions and 3 for the diagonal matrix.
All hidden vectors are of size 128.
Moreover,
we train our \mname~on all objects with 25 epochs
both separately and jointly,
as analyzed in the \Secref{sec:abl}.
The length of predictions $T$ in \Eqref{eq:loss} increases after every epoch,
which is capped at $T=16$.
We adopt the Adam optimizer
with an initial learning rate of 0.0008, which starts to decrease with a factor of 0.5 after 16 epochs.
All experiments are conducted on four NVIDIA A800 GPUs with a batch size of 4,
taking 22 hours to converge during training.

\noindent\textbf{DreamGaussian4D.}
We make sure to use the same amount of images to train DreamGaussian4D (DG4D) and other models.
We vary the hyperparameters for training,
including the number of iterations, the learning rate, the number of sampled views for score distillation sampling, etc.
We also vary the black background and white background to find the best settings for training.
Notice that the dynamic results reconstructed by DG4D tend to be slightly more blurred than the reference images,
which aligns with the fact in the official website.

\noindent\textbf{PhysDreamer.}
We adopt 768 sub-steps between adjacent video frames with a duration of $4.34 \times 10^{-5}$ seconds per sub-step.
Since PD's performance is \emph{highly dependent on hyperparameters and initialization configurations},
we try our best to find the best hyperparameters and initial values for each domain. We also try to change the grid size that maximizes the performance during training and test.
Notice that though we vary the initial young's modulus from 1.0 to $5\times10^{8}$ to find the best value to represent the stiffness, PD cannot support objects with high stiffness and swaying frequency, such as the synthetic ``Bunny'' and real-world ``Duck''.
Moreover, we observe that PD tends to predict rapid dissipation, which aligns with the phenomenon in the official website.

\input{tab_fig/app_qual}
\subsection{Rendering Results from Different Views}
In \Figref{fig:app_qual},
we exhibit more rendering results on different views.
\mname~achieves superior performance regardless of the view directions of the cameras,
suggesting the effectiveness in capturing the underlying physics laws.

\input{tab_fig/app_ssim}
\input{tab_fig/app_lpips}
\subsection{SSIM and LPIPS Evaluations}
In \Tabref{tbl:app_ssim} and \Tabref{tbl:app_lpips},
we report the SSIM for structural similarities and LPIPS for perceptual evaluations respectively.
Our \mname~achieves superior performance comparing with baselines.

\input{tab_fig/app_interact}
\subsection{Interactive Applications}
As shown in \Figref{fig:app_interact},
our method supports interactive dynamics by applying customized external forces $f$ on selected Gaussian kernels.
In practice,
we convert the forces as displacements of Gaussian kernels' positions to adapt to our simulation pipeline in \Eqref{eq:iter} as follows:
\begin{eqnarray}
    \Delta \vx &=& \frac{1}{2}\frac{f}{m}t^2,\\
    \mathcal{G}_{t+1} &=& \psi(\mathcal{G}_t+\Delta \vx, \mathcal{G}_{t-1}),
\end{eqnarray}
where $t$ is the time interval between two simulation steps and $m$ is the mass of the kernels as illustrated in \Secref{sec:hierarchical}.

%% file: tab_fig/app_symbols.tex
\setlength{\tabcolsep}{3pt}
\captionof{table}{
We list the mathematical symbols mentioned in the main manuscripts and supplementary materials as follows.
}
\label{tbl:app_symbols}
\begin{center} \small 
\begin{tabular}{cl}
\toprule
\bf Symbols	&\bf Descriptions\\\cmidrule(lr{.75em}){1-2}
$\mathcal{G}$    & Set of Gaussian kernels.\\
$\vx_k$ 		& The position of the $k$-th kernel.\\
$\bm{\sigma}_k, \bm{\Sigma}_k$ & The covariance of the $k$-th kernel in deformed space and material space respectively. \\
$\vc_k$ & The color of the $k$-th kernel.\\
$\alpha_k$ & The opacity of the $k$-th kernel.\\
$\rho_k$ & The trainable density of $k$-th kernel.\\
$\va_k$ & The trainable vector for attributes of $k$-th kernel.\\
$t$ & The timestamp. \\
$\psi_\theta(\cdot)$ & \mname, a neural network parameterized by $\theta$, designed to predict the future states of Gaussian kernels.\\
$I^o_t$ & The ground truth image at time $t$ for view direction $o$. \\
$\mX_k$ & The position of $k$-th kernel in material space.\\
$\vx_r, \mX_r$ & The position of the static kernel, which can be the root of the flower.\\
$\mF_k$ & The deformation gradient for $k$-th kernel to compute the deformed states at any timestamp.\\
$\mU_k, \mV_k$ & The components of ``Polar SVD'', representing the rotations. \\
$\mR_k$ & The rotation matrix computed by $\mU_k, \mV_k$, representing the closest rotation to the deformation gradient.\\ 
$\bm{\Lambda}, \bm{\lambda}$ & The diagonal matrix in ``Polar SVD'' decompositions and the 3D vector representing the singular values respectively.\\
$p,q,r$ & Non-negative real numbers.\\
$\vd$ & The 3D vector of view direction.\\
$g(\cdot)$ & The function to compute Gaussian kernel's color given view direction $\vd$.\\
$m_k, V_k$ & The mass and volume for $k$-th kernel.\\
$m_{c_l}, V_{c_l}$ & The mass and volume for $c_l$-th Center of Mass System.\\
$c_l$ & The index indicating the Center of Mass System at $l$-th level of hierarchical structure.\\
$L$ & Total number of hierarchical level.\\
$\hat{\vx}_k^{h-1}, \hat{\vx}_{c_h}^{h-1}$ & The predicted positions given $h$-th level's simulation for the center of $k$-th kernel and $c_h$-th Center of Mass System respectively.\\
$\hat{\bm{\sigma}}_k^{h-1}, \hat{\vc_k^{h-1}}$ & The predicted covariance and color for $k$-th kernel respectively.\\
$\mathcal{F}(\cdots)$ & The function to render Gaussian kernels for given view direction.\\
$G_k(\cdot)$ & The Gaussian function for $k$-th Gaussian kernel.\\
\bottomrule
\end{tabular}
\end{center}

%% file: tab_fig/app_radii.tex
\begin{table}[t]
    \caption{
    \small{
    The radii used for clustering kernels to construct the Hierarchical Center of Mass Systems at level $1$ and level $2$.
    }}
    \label{tbl:app_radii}
    \setlength{\tabcolsep}{1.8pt}
    \begin{center}
    \footnotesize 
    \begin{tabular}{lccccc}
    \toprule
    \bf Hierarchical Level	    & \bf Mothorchids 	    & \bf Carnation   & \bf Pudding	& \bf Duck  & \bf Bunny
    \\ \midrule
    l=1	& 0.04 	& 0.03  & 0.04    & 0.035   & 0.03\\
    l=2	& 0.5 	& 0.3    & 0.4    & 0.35 &0.3\\
    \bottomrule
    \end{tabular}
    \end{center}
\end{table}

%% file: tab_fig/app_color.tex
\begin{table*}[t]
    \caption{
    \small{
    The average color values normalized between 0 to 1 for foreground objects. Notice that the black background during training has a value of $(0,0,0)$, leading to larger contrast of color with the ``Bunny''.
    We also report the maximum value of $\ell$2 errors on sampled data.
    }}
    \label{tbl:app_contrast}
    \setlength{\tabcolsep}{2.8pt}
    \begin{center}
    \small
    \begin{tabular}{lccccc}
    \toprule
    & \bf Mothorchids	    & \bf Carnation 	    & \bf Pudding	& \bf Duck   & \bf Bunny
    \\ \midrule
    Avg Color	& (0.22, 0.14, 0.14) 	& (0.21, 0.23, 0.18)    & (0.24, 0.15, 0.06)  & (0.35, 0.31, 0.17)  & (0.78, 0.73, 0.67) \\
    Avg Color's $\ell$2	& 0.30 	& 0.37    & 0.29    & 0.50  & 1.26\\
    Max $\ell$2 Loss	& 0.67 	& 0.91    & 0.87   & 0.95  & 1.63\\
    \bottomrule
    \end{tabular}
    \end{center}
\end{table*}

%% file: tab_fig/app_qual.tex
\begin{figure*}[!h]
    \vspace{-4mm}
        \begin{center}
            \includegraphics[width=0.95\textwidth]{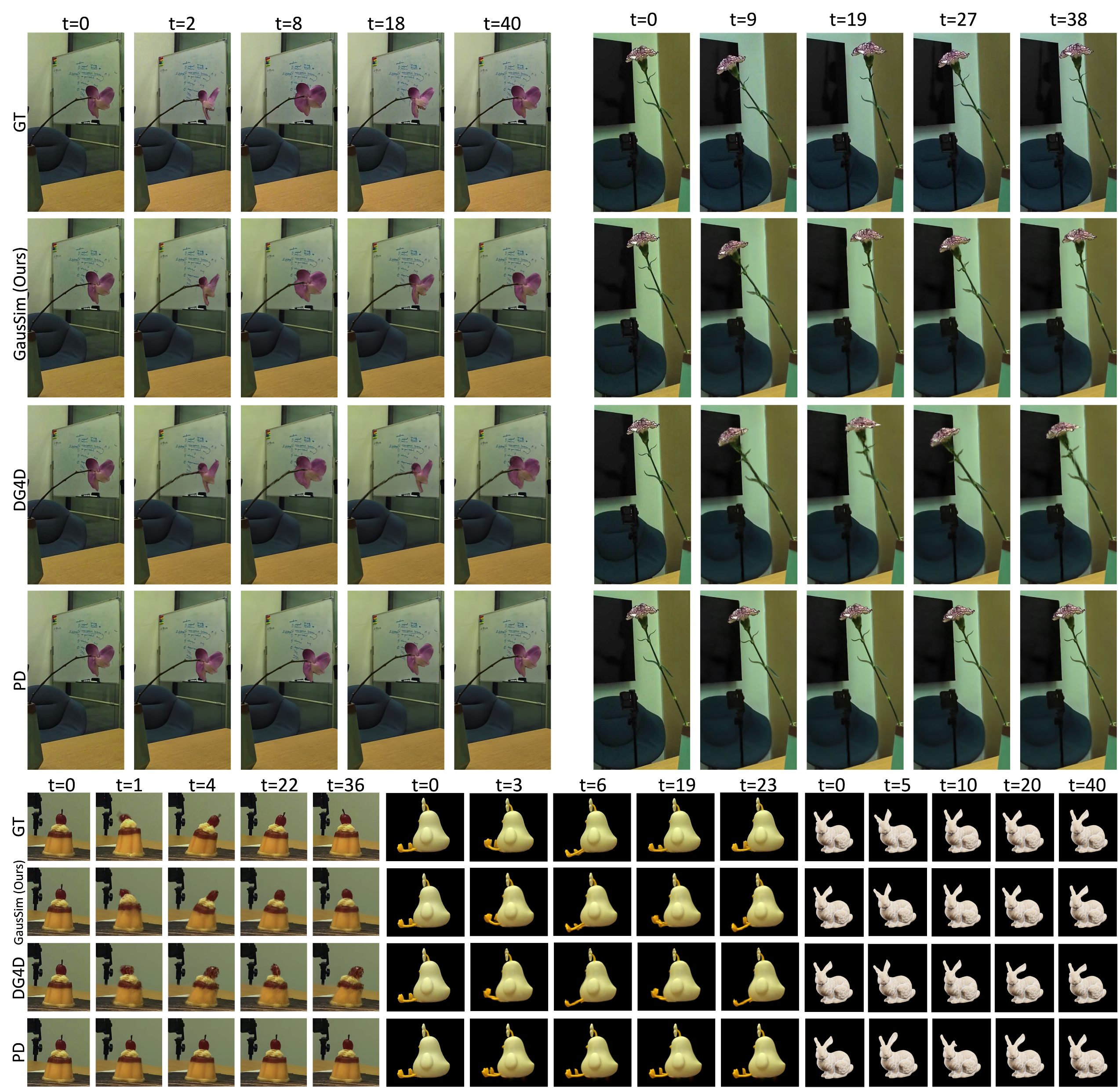}
        \end{center}
        \vspace{-4mm}
        \caption{\small{
        More qualitative results from different views.
        \mname~exhibits realistic deformations even with challenging initial conditions,
        indicating the effectiveness of our methods.
        }
        }
        \label{fig:app_qual}
\end{figure*}

%% file: tab_fig/app_ssim.tex
\begin{table*}[ht]
\setlength{\tabcolsep}{3pt}
\caption{
\small{
SSIM ($1\times10^{-1}$) results on our real dataset consisting of ``Mothorchids'', ``Carnation'',  ``Pudding'', and ``Duck'',
as well as synthetic ``Bunny''.
For the unseen frames,
we report the evaluations on 83 frames from $t=16$ to $t=99$ on ``Mothorchids'', ``Carnation'', and ``Pudding'',
and 34 frames from $t=16$ to $t=49$ on ``Duck'' and ``Bunny''.
Our \mname~achieves superior performance in all cases.}
}
\label{tbl:app_ssim}
\begin{center} \footnotesize
\begin{tabular}{lcccccccccc}
\toprule
\multirow{2}{*}{\bf SSIM ($\uparrow$)}	&\multicolumn{2}{c}{\bf{Mothorchids}}    & \multicolumn{2}{c}{\bf Carnation} & \multicolumn{2}{c}{\bf Pudding} & \multicolumn{2}{c}{\bf Duck} & \multicolumn{2}{c}{\bf Bunny} \\ \cmidrule(lr{.75em}){2-3}\cmidrule(lr{.75em}){4-5}\cmidrule(lr{.75em}){6-7}\cmidrule(lr{.75em}){8-9}\cmidrule(lr{.75em}){10-11}
						&Seen		&Unseen 83 		&Seen		         &Unseen 83		&Seen 		&Unseen 83          &Seen		&Unseen 34 & Seen & Unseen 34
\\ \cmidrule(lr{.75em}){1-1}\cmidrule(lr{.75em}){2-3}\cmidrule(lr{.75em}){4-5}\cmidrule(lr{.75em}){6-7}\cmidrule(lr{.75em}){8-9}\cmidrule(lr{.75em}){10-11}
\bf DG4D \cite{DBLP:journals/corr/abs-2312-17142}		    & 9.89$\pm$0.02		& 9.82$\pm$0.03	        & 9.80$\pm$0.01    &9.65$\pm$0.03    &9.91$\pm$0.00		& 9.88$\pm$0.01   &9.93$\pm$0.00  & 9.92$\pm$0.00 & 9.96$\pm$0.00 & 9.91$\pm$0.00\\
\bf PD \cite{DBLP:journals/corr/abs-2404-13026}		    & 9.76$\pm$0.02		& 9.86$\pm$0.02	        & 9.71$\pm$0.02    & 9.73$\pm$0.02    &9.91$\pm$0.00		&9.92$\pm$0.00   &9.91$\pm$0.00  & 9.92$\pm$0.01    &9.89$\pm$0.00  &9.86$\pm$0.00\\\midrule
\bf \mname(Ours) 		& \textbf{9.90$\pm$0.01}		& \textbf{9.90$\pm$0.01}	        & \textbf{9.81$\pm$0.01}  &\textbf{9.74$\pm$0.02}	&\textbf{9.95$\pm$0.00}  &\textbf{9.95$\pm$0.00}		        &\textbf{9.94$\pm$0.01}	&\textbf{9.95$\pm$0.01} & \textbf{9.97$\pm$0.00}  & \textbf{9.93$\pm$0.01}\\
\bottomrule
\end{tabular}
\end{center}
\end{table*}

%% file: tab_fig/app_lpips.tex
\begin{table*}[ht]
\setlength{\tabcolsep}{3pt}
\caption{
\small{
LPIPS ($1\times10^{-2}$) results on our real dataset consisting of ``Mothorchids'', ``Carnation'',  ``Pudding'', and ``Duck'',
as well as synthetic ``Bunny''.
For the unseen frames,
we report the evaluations on 83 frames from $t=16$ to $t=99$ on ``Mothorchids'', ``Carnation'', and ``Pudding'',
and 34 frames from $t=16$ to $t=49$ on ``Duck'' and ``Bunny''.
Our \mname~achieves superior performance in all cases.}
}
\label{tbl:app_lpips}
\begin{center} \footnotesize
\begin{tabular}{lcccccccccc}
\toprule
\multirow{2}{*}{\bf LPIPS ($\downarrow$)}	&\multicolumn{2}{c}{\bf{Mothorchids}}    & \multicolumn{2}{c}{\bf Carnation} & \multicolumn{2}{c}{\bf Pudding} & \multicolumn{2}{c}{\bf Duck} & \multicolumn{2}{c}{\bf Bunny} \\ \cmidrule(lr{.75em}){2-3}\cmidrule(lr{.75em}){4-5}\cmidrule(lr{.75em}){6-7}\cmidrule(lr{.75em}){8-9}\cmidrule(lr{.75em}){10-11}
						&Seen		&Unseen 83 		&Seen		         &Unseen 83		&Seen 		&Unseen 83          &Seen		&Unseen 34 & Seen & Unseen 34
\\ \cmidrule(lr{.75em}){1-1}\cmidrule(lr{.75em}){2-3}\cmidrule(lr{.75em}){4-5}\cmidrule(lr{.75em}){6-7}\cmidrule(lr{.75em}){8-9}\cmidrule(lr{.75em}){10-11}
\bf DG4D \cite{DBLP:journals/corr/abs-2312-17142}		    & 2.54$\pm$0.21		& 3.41$\pm$0.54	        & 4.00$\pm$0.19    &7.00$\pm$0.61    &1.42$\pm$0.03		&1.72$\pm$0.05   &1.52$\pm$0.13  & 1.82$\pm$0.14 & 0.27$\pm$0.02 & 0.69$\pm$0.01\\
\bf PD \cite{DBLP:journals/corr/abs-2404-13026}		    & 3.62$\pm$0.24		& 2.77$\pm$0.11	        & 4.56$\pm$0.27    & 5.66$\pm$0.54    &1.38$\pm$0.02		&1.23$\pm$0.04   &1.41$\pm$0.15  & 1.32$\pm$0.17    &1.67$\pm$0.19  &0.89$\pm$0.13\\\midrule
\bf \mname(Ours) 		& \textbf{2.46$\pm$0.10}		& \textbf{2.47$\pm$0.07}	        & \textbf{3.91$\pm$0.16}  &\textbf{5.10$\pm$0.58}	&\textbf{1.13$\pm$0.03}  &\textbf{1.17$\pm$0.04}		        &\textbf{1.18$\pm$0.18}	&\textbf{1.23$\pm$0.19} & \textbf{0.21$\pm$0.02}  & \textbf{0.49$\pm$0.08}\\
\bottomrule
\end{tabular}
\end{center}
\end{table*}

%% file: tab_fig/app_interact.tex
\begin{figure*}[!h]
    \vspace{-4mm}
        \begin{center}
            \includegraphics[width=0.95\textwidth]{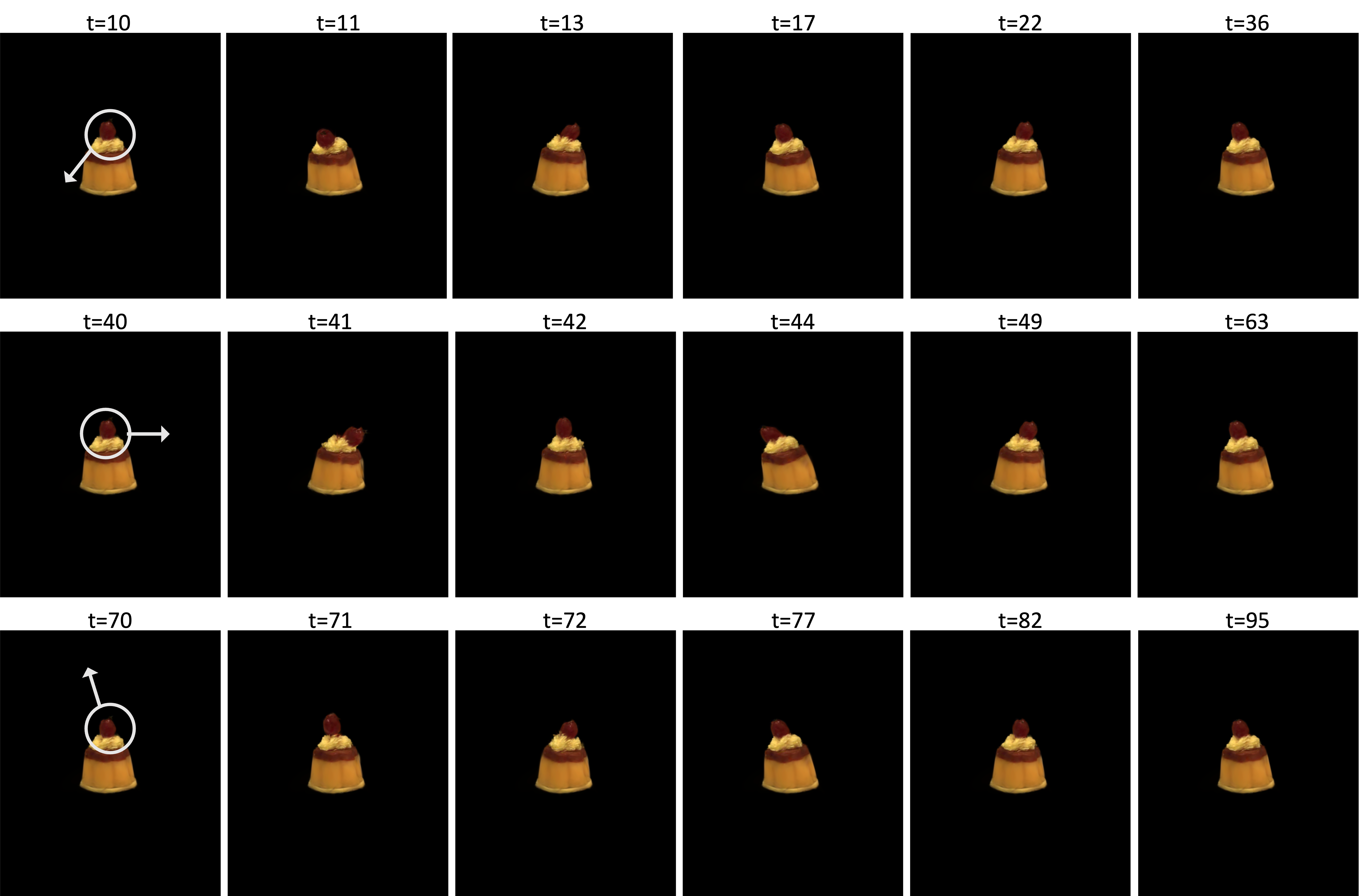}
        \end{center}
        \vspace{-4mm}
        \caption{\small{
        Interactive dynamics on pudding by \mname.
        We randomly drag the pudding along the arrows' directions.
        \mname~can vividly simulate the dynamics given external forces,
        suggesting the effectiveness and robustness of our method.
        }
        }
        \vspace{-4mm}
        \label{fig:app_interact}
\end{figure*}